\documentclass[10pt,twocolumn,letterpaper]{article}

\usepackage{cvpr}
\usepackage{times}
\usepackage{epsfig}
\usepackage{graphicx}
\usepackage{amsmath}
\usepackage{amssymb}
\usepackage{subfigure}
\usepackage{url}

\usepackage[pagebackref=true,breaklinks=true,letterpaper=true,colorlinks,bookmarks=false]{hyperref}

\cvprfinalcopy 

\def\cvprPaperID{2503} 

\ifcvprfinal\fi
 
\begin{document}
	
	\title{Age Progression/Regression by Conditional Adversarial Autoencoder}
	
	\author{Zhifei~Zhang\thanks{with equal contribution.}~,~~Yang~Song$^*$,~~Hairong~Qi\\
		The University of Tennessee, Knoxville, TN, USA\\
		{\tt\small \{zzhang61,~ysong18,~hqi\}@utk.edu}
	}

	\maketitle

	\begin{abstract}
		``If I provide you a face image of mine (without telling you the actual age when I took the picture) and a large amount of face images that I crawled (containing labeled faces of different ages but not necessarily paired), can you show me what I would look like when I am 80 or what I was like when I was 5?'' The answer is probably a ``No.'' Most existing face aging works attempt to learn the transformation between age groups and thus would require the paired samples as well as the labeled query image. In this paper, we look at the problem from a generative modeling perspective such that no paired samples is required. In addition, given an unlabeled image, the generative model can directly produce the image with desired age attribute. We propose a conditional adversarial autoencoder (CAAE) that learns a face manifold, traversing on which smooth age progression and regression can be realized simultaneously. In CAAE, the face is first mapped to a latent vector through a convolutional encoder, and then the vector is projected to the face manifold conditional on age through a deconvolutional generator. The latent vector preserves personalized face features (i.e., personality) and the age condition controls progression vs. regression. Two adversarial networks are imposed on the encoder and generator, respectively, forcing to generate more photo-realistic faces. Experimental results demonstrate the appealing performance and flexibility of the proposed framework by comparing with the state-of-the-art and ground truth. 
	\end{abstract}
	
	
	\begin{figure}[t]
		\centering
		\includegraphics[width=1\columnwidth]{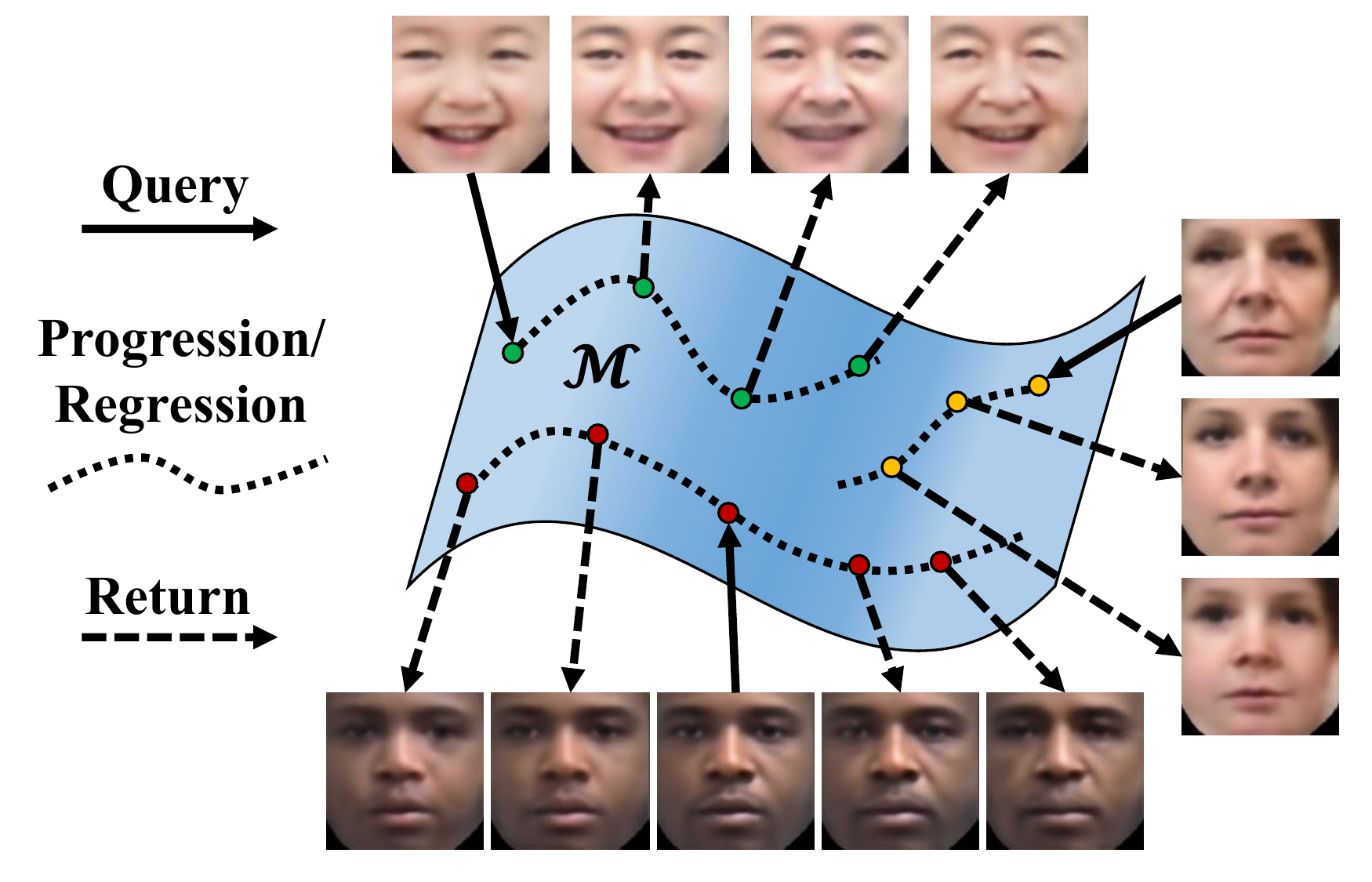}
		\caption{We assume the face images lie on a manifold ($\mathcal{M}$) , and images are clustered according to their ages and personality by a different direction. Given a query image, it will first projected to the manifold, and then after the smooth transformation on the manifold, the corresponding images will be projected back with aging patterns.}
		\label{fig:manifold}
	\end{figure}
	
\section{Introduction}
Face age progression (i.e., prediction of future looks) and regression (i.e., estimation of previous looks), also referred to as face aging and rejuvenation, aims to render face images with or without the ``aging'' effect but still preserve personalized features of the face (i.e., personality). It has tremendous impact to a wide-range of applications, \eg, face prediction of wanted/missing person, age-invariant verification, entertainment, etc. The area has been attracting a lot of research interests despite the extreme challenge in the problem itself. Most of the challenges come from the rigid requirement to the training and testing datasets, as well as the large variation presented in the face image in terms of expression, pose, resolution, illumination, and occlusion. The rigid requirement on the dataset refers to the fact that most existing works require the availability of “paired” samples, i.e., face images of the same person at different ages, and some even require paired samples over a long range of age span, which is very difficult to collect. For example, the largest aging dataset ``Morph"~\cite{kemelmacher2014illumination} only captured images with an average time span of 164 days for each individual. In addition, existing works also require the query image to be labeled with the true age, which can be inconvenient from time to time. 
Given the training data, existing works normally divide them into different age groups and learn a transformation between the groups, therefore, the query image has to be labeled in order to correctly position the image. 

Although age progression and regression are equally important, most existing works focus on age progression. Very 
few works can achieve good performance of face rejuvenating, especially for rendering baby face from an adult because they are mainly surface-based modeling which simply remove the texture from a given image~\cite{liu2004image,lanitis2002toward,fu2006m}. On the other hand, researchers have made great progress on age progression. For example, the physical model-based methods~\cite{tazoe2012facial,suo2010compositional,lanitis2002toward,ramanathan2006modeling} parametrically model biological facial change with age, \eg, muscle, wrinkle, skin, etc. However, they suffer from complex modeling, the requirement of sufficient dataset to cover long time span, and are computationally expensive; the prototype-based methods~\cite{tiddeman2001prototyping,kemelmacher2014illumination,shu2015personalized,wangrecurrent} tend to divide training data into different age groups and learn a transformation between groups. However, some can preserve personality but induce severe ghosting artifacts, others smooth out the ghosting effect but lose personality, while most relaxed the requirement of paired images over long time span, and the aging pattern can be learned between two adjacent age groups. Nonetheless, they still need paired samples over short time span. 

In this paper, we investigate the age progression/regression problem from the perspective of generative modeling. 
The rapid development of generative adversarial networks (GANs) has shown impressive results in face image generation~\cite{Alireza,zhao2016energy,radford2015unsupervised,liu2016coupled}. 
In this paper, we assume that the face images lie on a high-dimensional manifold as shown in Fig.\ref{fig:manifold}. Given a query face, we could find the corresponding point (face) on the manifold. Stepping along the direction of age changing, we will obtain the face images of different ages while preserving personality. We propose a conditional adversarial autoencoder (CAAE)\footnote{Bitbucket: {\scriptsize\url{https://bitbucket.org/aicip/face-aging-caae}}\\ \indent\hspace{4.5 px} Github: \scriptsize\url{https://zzutk.github.io/Face-Aging-CAAE}} network to learn the face manifold. By controlling the age attribute, it will be flexible to achieve age progression and regression at the same time. Because it is difficult to directly manipulate on the high-dimensional manifold, the face is first mapped to a latent vector through a convolutional encoder, and then the vector is projected to the face manifold conditional on age through a deconvolutional generator.
 Two adversarial networks are imposed on the encoder and generator, respectively, forcing to generate more photo-realistic faces. 

The benefit of the proposed CAAE can be summarized from four aspects. First, the novel network architecture achieves both age progression and regression while generating photo-realistic face images. Second, we deviate from the popular group-based learning, thus not requiring paired samples in the training data or labeled face in the test data, making the proposed framework much more flexible and general. Third, the disentanglement of age and personality in the latent vector space helps preserving personality while avoiding the ghosting artifacts. Finally, CAAE is robust against variations in pose, expression, and occlusion.

\section{Related Work}
\subsection{Age Progression and Regression}

	In recent years, the study on face age progression has been very popular, with approaches mainly falling into two categories, physical model-based and prototype-based. Physical model-based methods
model the biological pattern and physical mechanisms of aging, \eg, the muscles~\cite{suo2012concatenational}, wrinkle~\cite{ramanathan2008modeling,suo2010compositional}, facial structure~\cite{ramanathan2006modeling,lanitis2002toward} etc. through either parametric or non-parametric learning. However, in order to better model the subtle aging mechanism, it will require a large face dataset with long age span (\eg, from 0 to 80 years old) of each individual, which is very difficult to collect. In addition, physical modeling-based approaches are computationally expensive.
	
	On the other hand, prototype-based approaches \cite{burt1995perception, kemelmacher2014illumination} often divide faces into groups by age, e.g., the average face of each group, as its prototype. Then, the difference between prototypes from two age groups is considered the aging pattern. However, the aged face generated from averaged prototype may lose the personality (\eg, wrinkles). To preserve the personality,~\cite{shu2015personalized} proposed a dictionary learning based method --- age pattern of each age group is learned into the corresponding sub-dictionary. A given face will be decomposed into two parts: age pattern and personal pattern. The age pattern was transited to the target age pattern through the sub-dictionaries, and then the aged face is generated by synthesizing the personal pattern and target age pattern. However, this approach presents serious ghosting artifacts. 
	The deep learning-based method \cite{wangrecurrent} represents the state-of-the-art, where RNN is applied on the coefficients of eigenfaces for age pattern transition. All prototype-based approaches perform the group-based learning which requires the true age of testing faces to localize the transition state which might not be convenient. In addition, these approaches only provide age progression from younger face to older ones. To achieve flexible bidirectional age changes, it may need to retrain the model inversely. 
	
	Face age regression, which predicts the rejuvenating results, is comparatively more challenging. Most age regression works so far \cite{liu2004image,fu2006m} are physical model-based, where the textures are simply removed based on the learned transformation over facial surfaces. Therefore, they cannot achieve photo-realistic results for baby face predictions. 
	
	\subsection{Generative Adversarial Network}
	
	Generating realistically appealing images is still challenging and has not achieved much success until the rapid advancement of the generative adversarial network (GAN). The original GAN work~\cite{goodfellow2014generative} introduced a novel framework for training generative models. It simultaneously trains two models: 1) the generative model $G$ captures the distribution of training samples and learns to generate new samples imitating the training, and 2) the discriminative model $D$ discriminates the generated samples from the training. $G$ and $D$ compete with each other using a min-max game as Eq.~\ref{eq:GAN}, where $z$ denotes a vector randomly sampled from certain distribution $p(\mathbf{z})$ (\eg, Gaussian or uniform), and the data distribution is $p_{data}(\mathbf{x})$, \ie, the training data $x\sim p_{data}(\mathbf{x})$. 
	\begin{equation}
	\centering
	\begin{split}
	\underset{G}{\min} \;\underset{D}{\max}\;&\mathbb{E}_{x\sim p_{data}(\mathbf{x})}[\log D(x)] + \\ 
	&\mathbb{E}_{z\sim p(\mathbf{z})}[\log(1-D(G(z)))]
	\end{split}
	\label{eq:GAN}
	\end{equation}
	The two parts, $G$ and $D$, are trained alternatively. 
	
	One of the biggest issues of GAN is that the training process is unstable, and the generated images are often noisy and incomprehensible. During the last two years, several approaches~\cite{radford2015unsupervised,mirza2014conditional,gregor2015draw,Chen2016InfoGAN,denton2015deep,im2016generating,Alireza} have been proposed to improve the original GAN from different perspectives. For example, DCGAN~\cite{radford2015unsupervised} adopted deconvolutional and convolutional neural networks to implement $G$ and $D$, respectively. It also provided empirical instruction on how to build a stable GAN, \eg, replacing the pooling by strides convolution and using batch normalization. CGAN~\cite{mirza2014conditional} modified GAN from unsupervised learning into semi-supervised learning by feeding the conditional variable (\eg, the class label) into the data. The low resolution of the generated image is another common drawback of GAN. \cite{denton2015deep,im2016generating} extended GAN into sequential or pyramid GANs to handle this problem, where the image is generated step by step, and each step utilizes the information from the previous step to further improve the image quality. 
	Some GAN-related works have shown visually impressive results of randomly drawing face images~\cite{yu2016ultra,Alireza,zhao2016energy,radford2015unsupervised,liu2016coupled}. 
	However, GAN generates images from random noise, thus the output image cannot be controlled. This is undesirable in age progression and regression, where we have to ensure the output face looks like the same person as queried. 

\section{Traversing on the Manifold}
\begin{figure}[t]
	\centering
	\includegraphics[width=\columnwidth]{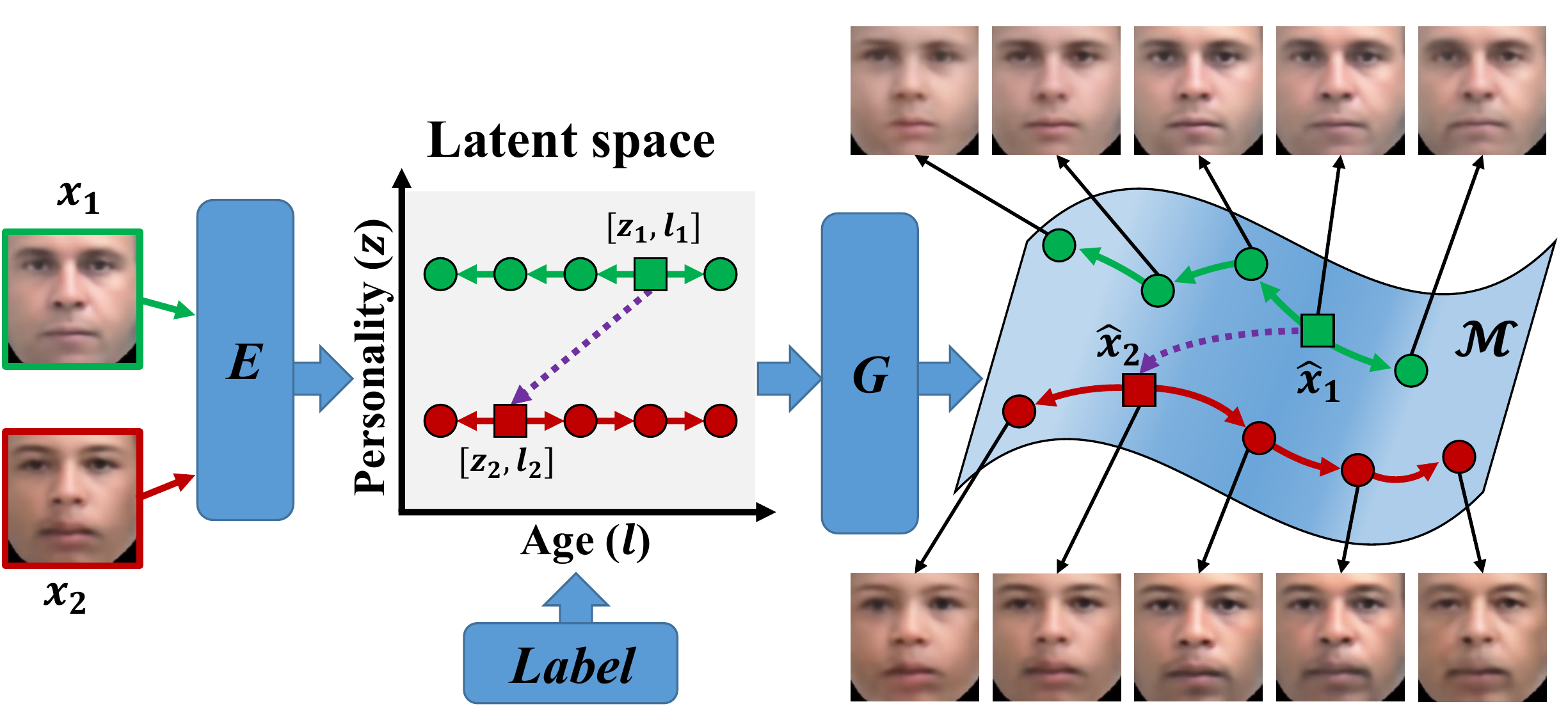}
	\caption{Illustration of traversing on the face manifold $\mathcal{M}$.
		The input faces $x_1$ and $x_2$ are encoded to $z_1$ and $z_2$ by an encoder $E$, which represents the personality. Concatenated by random age labels $l_1$ and $l_2$, the latent vectors $[z_1,l_1]$ and $[z_2,l_2]$ are constructed as denoted by the rectangular points. The colors indicate correspondence of personality. Arrows and circle points denote the traversing direction and steps, respectively. Solid arrows direct traversing along the age axis while preserving the personality. The dotted arrow performs a traversing across both the age and personality axes. The traversing in the latent space is mapped to the face manifold $\mathcal{M}$ by a generator $G$, as illustrated by the points and arrows with corresponding markers and colors. Each point on $\mathcal{M}$ is a face image, thus achieving age progression and regression.}
	\label{fig:traversing}
\end{figure}

We assume the face images lie on a high-dimensional manifold, on which traversing along certain direction could achieve age progression/regression while preserving the personality. This assumption will be demonstrated experimentally in Sec.~\ref{subsec:objective}. However, modeling the high-dimensional manifold is complicated, and it is difficult to directly manipulate (traversing) on the manifold. 
Therefore, we will learn a mapping between the manifold and a lower-dimensional space, referred to as the \textit{latent space}, which is easier to manipulate. 
As illustrated in Fig.~\ref{fig:traversing}, faces $x_1$ and $x_2$ are mapped to the latent space by $E$ (\ie, an encoder), which extracts the personal features $z_1$ and $z_2$, respectively. Concatenating with the age labels $l_1$ and $l_2$, two points are generated in the latent space, namely $[z_1, l_1]$ and $[z_2,l_2]$. Note that the personality $z$ and age $l$ are disentangled in the latent space, thus we could simply modify age while preserving the personality. Starting from the red rectangular point $[z_2,l_2]$ (corresponding to $x_2$) and evenly stepping bidirectionally along the age axis (as shown by the solid red arrows), we could obtain a series of new points (red circle points). Through another mapping $G$ (\ie. a generator), those points are mapped to the manifold $\mathcal{M}$ -- generating a series of face images, which will present the age progression/regression of $x_2$. By the same token, the green points and arrows demonstrate the age progressing/regression of $x_1$ based on the learned manifold and the mappings. If we move the point along the dotted arrow in the latent space, both personality and age will be changed as reflected on $\mathcal{M}$. We will learn the mappings $E$ and $G$ to ensure the generated faces lie on the manifold, which indicates that the generated faces are realistic and plausible for a given age.          

\begin{figure*}[t]
	\centering
	\includegraphics[width=2\columnwidth]{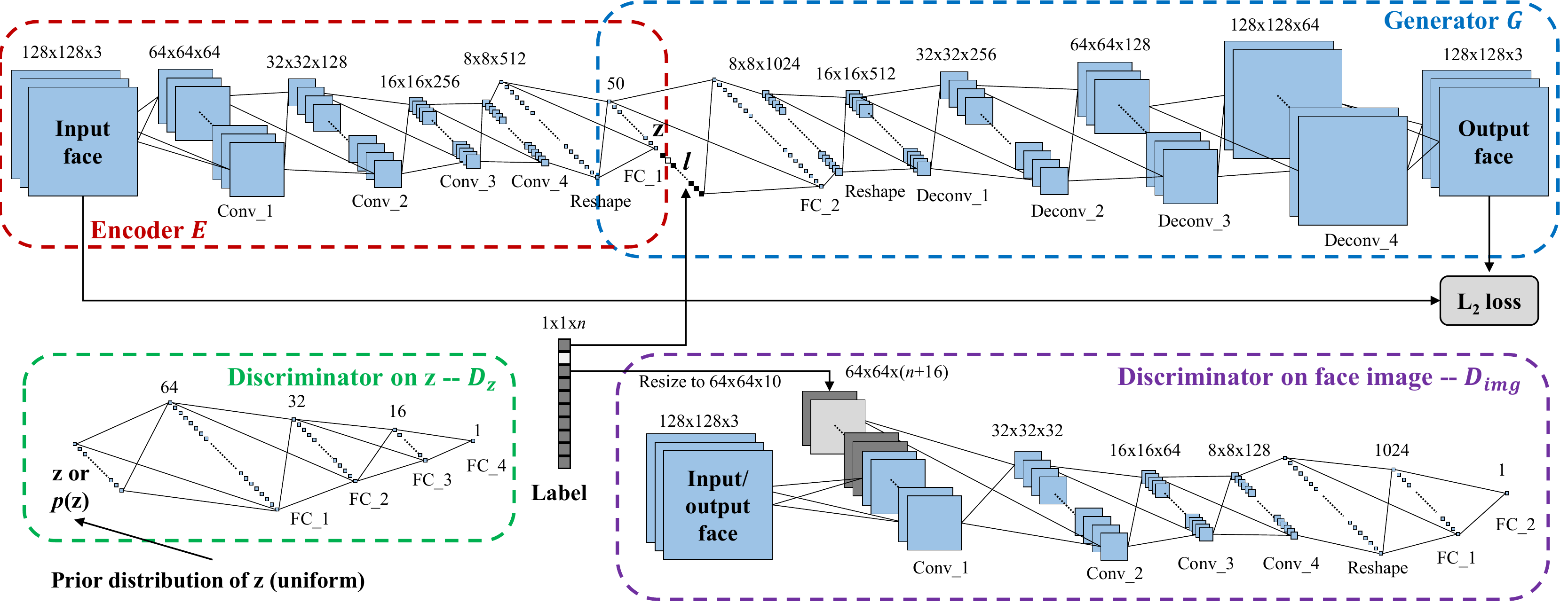}
	\caption{Structure of the proposed CAAE network for age progression/regression. 
		The encoder $E$ maps the input face to a vector $z$ (personality). Concatenating the label $l$ (age) to $z$, the new latent vector $[z,l]$ is fed to the generator $G$. Both the encoder and the generator are updated based on the $L_2$ loss between the input and output faces. The discriminator $D_z$ imposes the uniform distribution on $z$, and the discriminator $D_{img}$ forces the output face to be photo-realistic and plausible for a given age label.}
	\label{fig:flow}
\end{figure*}

\section{Approach}
In this section, we first present the pipeline of the proposed conditional adversarial autoencoder (CAAE) network (Sec.~\ref{subsec:CAAE}) that learns the face manifold (Sec.~\ref{subsec:objective}). The CAAE incorporates two discriminator networks, which are the key to generating more realistic faces. Sections~\ref{subsec:Dz} and \ref{subsec:DG} demonstrate their effectiveness, respectively. Finally, Section~\ref{subsec:diff} discusses the difference of the proposed CAAE from other generative models. 

\subsection{Conditional Adversarial Autoencoder}
\label{subsec:CAAE}
The detailed structure of the propose CAAE network is shown in Fig.~\ref{fig:flow}. The input and output face images are $128\times128$ RGB images $x\in\mathbb{R}^{128\times128\times3}$. A convolutional neural network is adopted as the encoder. The convolution of stride 2 is employed instead of pooling (\eg, max pooling) because strided convolution is fully differentiable and allows the network to learn its own spacial downsampling~\cite{radford2015unsupervised}. The output of encoder $E(x)=z$ preserves the high-level personal feature of the input face $x$. The output face conditioned on certain age can be expressed by $G(z,l)=\hat{x}$, where $l$ denotes the one-hot age label. 
Unlike existing GAN-related works, we incorporate an encoder to avoid random sampling of $z$ because we need to generate a face with specific personality which is incorporated in $z$. In addition, two discriminator networks are imposed on $E$ and $G$, respectively. The $D_z$ regularizes $z$ to be uniform distributed, smoothing the age transformation. The $D_{img}$ forces $G$ to generate photo-realistic and plausible faces for arbitrary $z$ and $l$. The effectiveness of the two discriminators will be further discussed in Secs.~\ref{subsec:Dz} and~\ref{subsec:DG}, respectively.


\subsection{Objective Function} 
\label{subsec:objective}
The real face images are supposed to lie on the face manifold $\mathcal{M}$, so the input face image $x\in\mathcal{M}$. The encoder $E$ maps the input face $x$ to a feature vector, \ie, $E(x)=z\in\mathbb{R}^n$, where $n$ is the dimension of the face feature. Given $z$ and conditioned on certain age label $l$, the generator $G$ generates the output face $\hat{x}=G(z,l)=G(E(x),l)$. Our goal is to ensure the output face $\hat{x}$ lies on the manifold while sharing the personality and age with the input face $x$ (during training). Therefore, the input and output faces are expected to be similar as expressed in Eq.~\ref{eq:loss_EG}, where $\mathcal{L}(\cdot,\cdot)$ denotes L$_2$ norm.
\begin{equation}
\centering
\underset{E,G}{\min}\; \mathcal{L}\left(x,G(E(x),l)\right)
\label{eq:loss_EG}
\end{equation}
Simultaneously, the uniform distribution is imposed on $z$ through $D_z$ -- the discriminator on $z$. We denote the distribution of the training data as $p_{data}(\mathbf{x})$, then the distribution of $z$ is $q(\mathbf{z}|\mathbf{x})$.
Assuming $p(\mathbf{z})$ is a prior distribution, and $z^*\sim p(\mathbf{z})$ denotes the random sampling process from $p(\mathbf{z})$. A min-max objective function can be used to train $E$ and $D_z$,
\begin{equation}
\centering
\begin{split}
\underset{E}{\min}\;\underset{D_z}{\max}\; & \mathbb{E}_{z^*\sim p(\mathbf{z})}\left[ \log D_z(z^*)  \right] + \\
~ & \mathbb{E}_{x\sim p_{data}(\mathbf{x})}\left[ \log (1-D_z(E(x)))  \right]
\end{split}
\label{eq:loss_EDz}
\end{equation}  
By the same token, the discriminator on face image, $D_{img}$, and $G$ with condition $l$  can be trained by
\begin{equation}
\centering
\begin{split}
\underset{G}{\min} & \; \underset{D_{img}} {\max} \;  \mathbb{E}_{x,l\sim p_{data}(\mathbf{x, l})}\left[ \log D_{img}(x,l)  \right] + \\
~&\mathbb{E}_{x,l\sim p_{data}(\mathbf{x, \l})}\left[ \log (1-D_{img}(G(E(x),l)))  \right]
\end{split}
\label{eq:loss_GDi}
\end{equation}
Finally the objective function becomes
\begin{equation}
\centering
\begin{split}
\underset{E,G}{\min}\; &\underset{D_z,D_{img}}{\max}  \lambda\mathcal{L}\left(x,G(E(x),l)\right) + \gamma TV(G(E(x),l))  \\
+	& \mathbb{E}_{z^*\sim p(\mathbf{z})}\left[ \log D_z(z^*)  \right] \\
+ & \mathbb{E}_{x\sim p_{data}(\mathbf{x})}\left[ \log (1-D_z(E(x)))  \right]  \\
+& \mathbb{E}_{x,l\sim p_{data}(\mathbf{x, l})}\left[ \log D_{img}(x,l)  \right]  \\
+& \mathbb{E}_{x,l\sim p_{data}(\mathbf{x, \l})}\left[ \log (1-D_{img}(G(E(x),l)))  \right],
\end{split}
\label{eq:objective}
\end{equation}
where $TV(\cdot)$ denotes the total variation which is effective in removing the ghosting artifacts. The coefficients $\lambda$ and $\gamma$ balance the smoothness and high resolution.

\begin{figure}[t]
	\centering
	\includegraphics[width=.9\columnwidth]{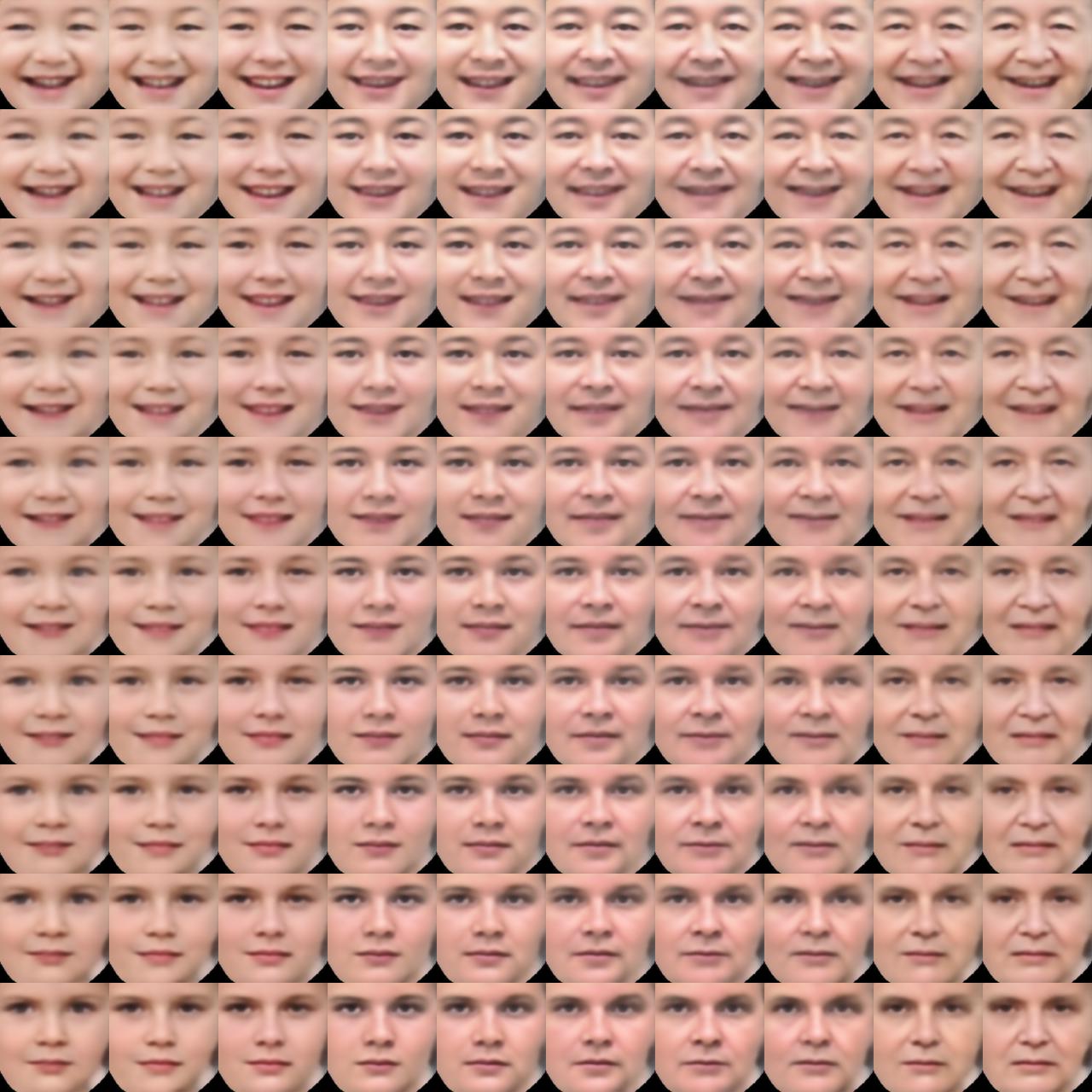}
	\caption{Illustration of the learned face manifold $\mathcal{M}$. The horizontal axis indicates the traversing of age, and the vertical axis indicates different personality.}
	\label{fig:demo}
\end{figure}

Note that the age label is resized and concatenated to the first convolutional layer of $D_{img}$ to make it discriminative on both age and human face. Sequentially updating the network by Eqs.~\ref{eq:loss_EG}, \ref{eq:loss_EDz}, and \ref{eq:loss_GDi}, we could finally learn the manifold $\mathcal{M}$ as illustrated in Fig.~\ref{fig:demo}.  

\subsection{Discriminator on $\boldsymbol{z}$}
\label{subsec:Dz}
\begin{figure}[t]
	\centering
	\includegraphics[width=.9\columnwidth]{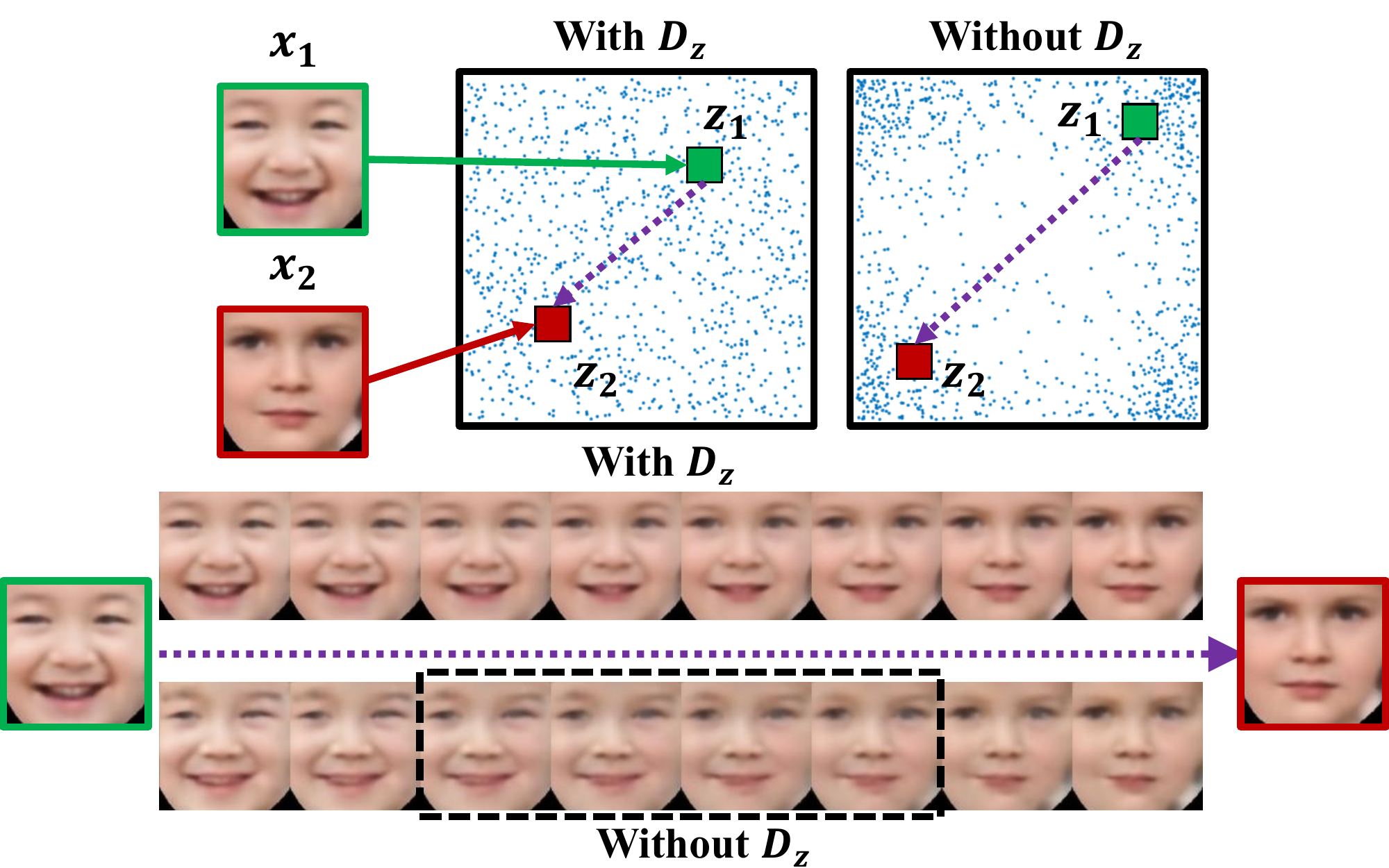}
	\caption{Effect of $D_z$, which forces $z$ to a uniform distribution. For simplicity, $z$ is illustrated in a 2-D space. Blue dots indicate $z$'s mapped from training faces through the encoder. With $D_z$, the distribution of $z$ will approach uniform. Otherwise, $z$ may present ``holes''. The rectangular points denote the corresponding $z$ mapped from the input faces $x_1$ and $x_2$, and the dotted arrow indicates the traversing from $z_1$ to $z_2$. The intermediate points along the traversing are supposed to generate a series of plausible morphing faces from $x_1$ to $x_2$. Without $D_z$, the learned $z$ presents a sparse distribution along the path of traversing, causing the generated face to look unreal. The series of figures at the bottom shows the traversing with and without $D_z$.}
	\label{fig:Dz}
\end{figure}

The discriminator on $z$, denoted by $D_z$, imposes a prior distribution (\eg, uniform distribution) on $z$. Specifically, $D_z$ aims to discriminate the $z$ generated by encoder $E$. Simultaneously, $E$ will be trained to generate $z$ that could fool $D_z$. Such adversarial process forces the distribution of the generated $z$ to gradually approach the prior. We use uniform distribution as the prior, forcing $z$ to evenly populate the latent space with no apparent ``holes''. As shown in Fig.~\ref{fig:Dz}, the generated $z$'s (depicted by blue dots in a 2-D space) present uniform distribution under the regularization of $D_z$, while the distribution of $z$ exhibits a ``hole'' without the application of $D_z$.    
Exhibition of the ``hole'' indicates that face images generated by interpolating between arbitrary $z$'s may not lie on the face manifold -- generating unrealistic faces. For example, given two faces $x_1$ and $x_2$ as shown in Fig.~\ref{fig:Dz}, we obtain the corresponding $z_1$ and $z_2$ by $E$ under the conditions with and without $D_z$, respectively. Interpolating between $z_1$ and $z_2$ (dotted arrows in Fig.~\ref{fig:Dz}), the generated faces are expected to show realistic and smooth morphing from $x_1$ to $x_2$ (bottom of Fig.~\ref{fig:Dz}). However, the morphing without $D_z$ actually presents distorted (unrealistic) faces in the middle (indicated by dashed box), which corresponds to the interpolated $z$'s passing through the ``hole''.       


\subsection{Discriminator on Face Images} 
\label{subsec:DG}
\begin{figure}[h]
	\centering
	\includegraphics[width=.7\columnwidth]{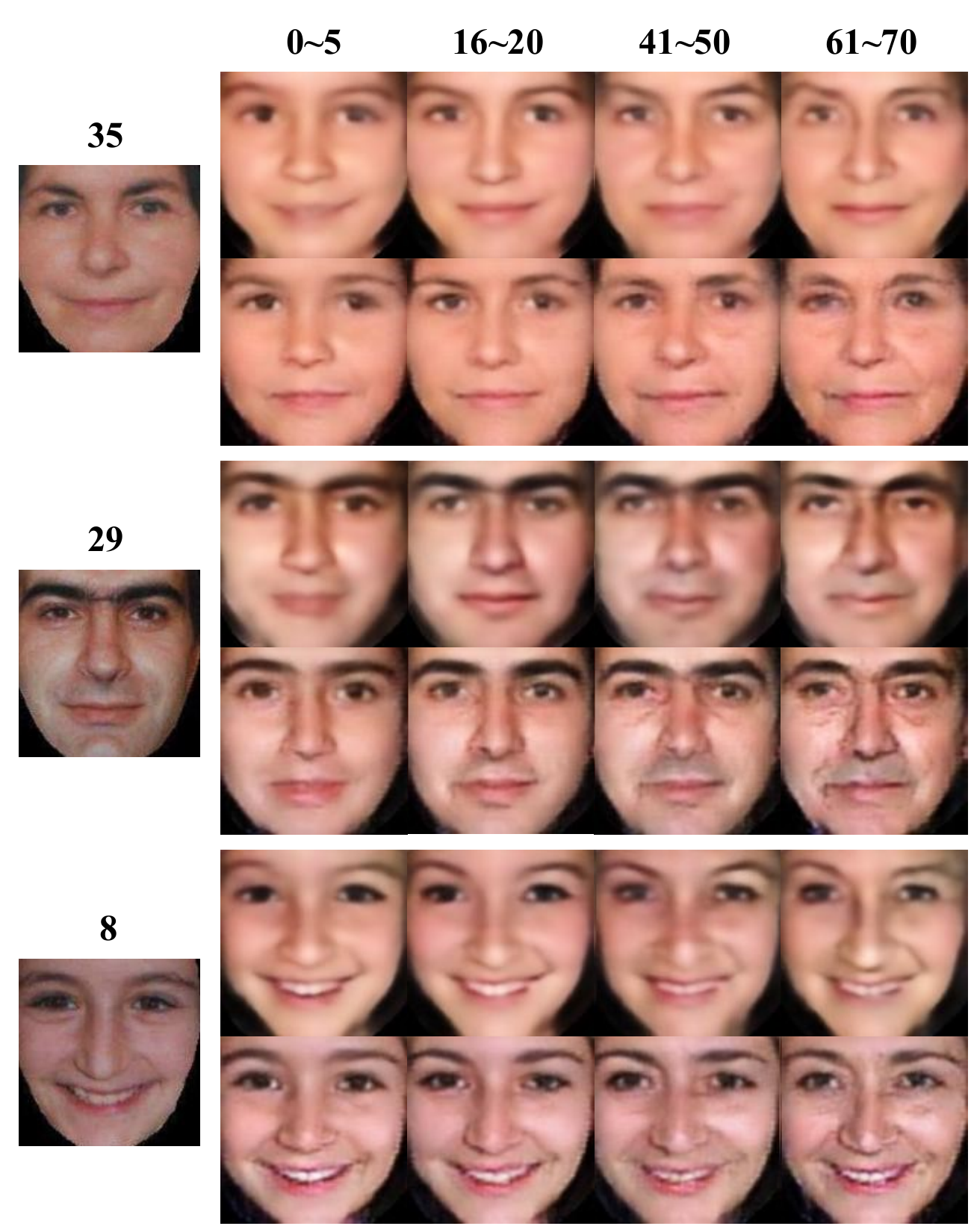}
	\caption{Effect of $D_{img}$, which forces the generated faces to be more realistic in aspects of age and resolution. The first column shows the original faces, and their true ages are marked on the top. The right four columns are generated faces through the proposed framework, without (the upper row) or with (the lower row) $D_{img}$. The generated faces fall in four age groups as indicated at the top of each column.}
	\label{fig:D_img}
\end{figure}

Inheriting the similar principle of GAN, the discriminator $D_{img}$ on face images forces the generator to yield more realistic faces. In addition, the age label is imposed on $D_{img}$ to make it discriminative against unnatural faces conditional on age. Although minimizing the distance between the input and output images as expressed in Eq.~\ref{eq:loss_EG} forces the output face to be close to the real ones, Eq.~\ref{eq:loss_EG} does not ensure the framework to generate plausible faces from those unsampled faces. For example, given a face that is unseen during training and a random age label, the pixel-wise loss could only make the framework generate a face close to the trained ones in a manner of interpolation, causing the generated face to be very blurred. The $D_{img}$ will discriminate the generated faces from real ones in aspects of reality, age, resolution, etc. Fig.~\ref{fig:D_img} demonstrates the effect of $D_{img}$.

Comparing the generated faces with and without $D_{img}$, it is obvious that $D_{img}$ assists the framework to generate more realistic faces. The outputs without $D_{img}$ could also present aging but the effect is not as obvious as that with $D_{img}$ because $D_{img}$ enhances the texture especially for older faces.   
\begin{figure*}[h]
	\centering
	\includegraphics[width=1.8\columnwidth]{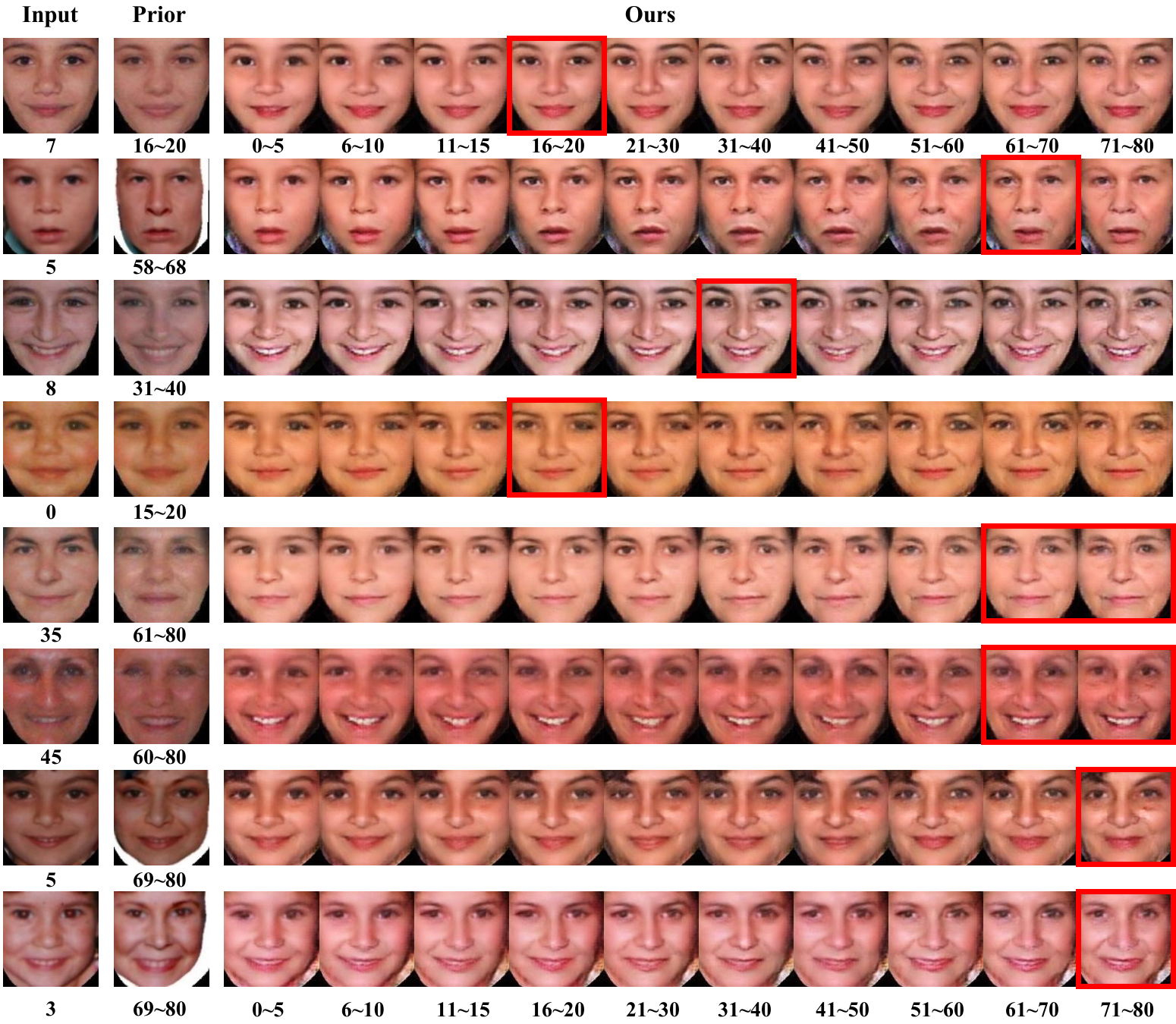}
	\caption{Comparison to prior works of face aging. The first column shows input faces, and second column are the best aged faces cited from prior works. The rest columns are our results from both age progression and regression. The red boxes indicate the comparable results to the prior works.}
	\label{fig:cmp2prior}
\end{figure*}

\subsection{Differences from Other Generative Networks}
\label{subsec:diff}
In this section, we comment on the similarity and difference of the proposed CAAE with other generative networks, including GAN~\cite{goodfellow2014generative}, variational autoencoder (VAE)~\cite{kingma2013auto}, and adversarial autoencoder (AAE)~\cite{Alireza}. 

\textbf{VAE vs. GAN}: 
VAE uses a recognition network to predict the posterior distribution over the latent variables, while GAN uses an adversarial training procedure to directly shape the output distribution of the network via back-propagation~\cite{Alireza}. Because VAE follows an encoding-decoding scheme, we can directly compare the generated images to the inputs, which is not possible when using a GAN. A downside of VAE is that it uses mean squared error instead of an adversarial network in image generation, so it tends to produce more blurry images~\cite{larsen2015autoencoding}.

\textbf{AAE vs. GAN and VAE}:
AAE can be treated as the combination of GAN and VAE, which maintains the autoencoder network like VAE but replaces the KL-divergence loss with an adversarial network like in GAN. Instead of generating images from random noise as in GAN, AAE utilizes the encoder part to learn the latent variables approximated on certain prior, making the style of generated images controllable.
In addition, AAE better captures the data manifold compared to VAE.

\textbf{CAAE vs. AAE}:
The proposed CAAE is more similar to AAE. The main difference from AAE is that the proposed CAAE imposes discriminators on the encoder and generator, respectively. The discriminator on encoder guarantees smooth transition in the latent space, and the discriminator on generator assists to generate photo-realistic face images. Therefore, CAAE would generate higher quality images than AAE as discussed in Sec.~\ref{subsec:DG}.

\section{Experimental Evaluation}
\label{sec:experiment}
In the section, we will first clarify the process of data collection (Sec.~\ref{subsec:exp_data}) and implementation of the proposed CAAE (Sec.~\ref{subsec:exp_implementation}). Then, both qualitative and quantitative comparisons with prior works and ground truth are performed in Sec.~\ref{subsec:exp_comparison}. Finally, the tolerance to occlusion and variation in pose and expression is illustrated in Sec.~\ref{subsec:exp_tolerance} .


\subsection{Data Collection}
\label{subsec:exp_data}
We first collect face images from the Morph dataset~\cite{kemelmacher2014illumination} and the CACD~\cite{chen14cross} dataset. The Morph dataset~\cite{kemelmacher2014illumination} is the largest with multiple ages of each individual, including 55,000 images of 13,000 subjects from 16 to 77 years old.
The CACD~\cite{chen14cross} dataset contains 13,446 images of 2,000 subjects. Because both datasets have limited images from newborn or very old faces, we crawl images from Bing and Google search engines based on the keywords, \eg, baby, boy, teenager, 15 years old, etc. Because the proposed approach does not require multiple faces from the same subject, we simply randomly choose around 3,000 images from the Morph and CACD dataset and crawl 7,670 images from the website. The age and gender of the crawled faces are estimated based on the image caption or the result from age estimator~\cite{levi2015age}. We divide the age into ten categories, \ie, 0--5, 6--10, 11--15, 16--20, 21--30, 31--40, 41--50, 51--60, 61--70, and 71--80. Therefore, we can use a one-hot vector of ten elements to indicate the age of each face during training. The final dataset consists of 10,670 face images with a uniform distribution on gender and age. We use the face detection algorithm with 68 landmarks~\cite{dlib} to crop out and align the faces, making the training more attainable. 

\subsection{Implementation of CAAE}
\label{subsec:exp_implementation}
We construct the network according to Fig.~\ref{fig:flow} with kernel size of $5 \times 5$. The pixel values of the input images are normalized to $[-1,1]$, and the output of $E$ (\ie, $z$) is also restricted to $[-1,1]$ by the hyperbolic tangent activation function. Then, the desired age label, the one-hot vector, is concatenated to $z$, constructing the input of $G$. To make fair concatenation, the elements of label is also confined to $[-1,1]$, where -1 corresponds to 0. Finally, the output is also in range $[-1,1]$ through the hyperbolic tangent function. Normalizing the input may make the training process converge faster. Note that we will not use the batch normalization for $E$ and $G$ because it blurs personal features and makes output faces drift far away from inputs in testing. However, the batch normalization will make the framework more stable if it is applied on $D_{img}$. 
All intermediate layers of each block (\ie, $E$, $G$, $D_z$, and $D_{img}$) use the ReLU activation function. 

In training, $\lambda=100$, $\gamma=10$, and the four blocks are updated alternatively with a mini-batch size of 100 through the stochastic gradient descent solver, ADAM~\cite{kingma2014adam} ($\alpha=0.0002$, $\beta_1=0.5$). Face and age pairs are fed to the network. After about 50 epochs, plausible generated faces can be obtained. During testing, only $E$ and $G$ are active. Given an input face without true age label, $E$ maps the image to $z$. Concatenating an arbitrary age label to $z$, $G$ will generate a photo-realistic face corresponding to the age and personality.


\subsection{Qualitative and Quantitative Comparison}
\label{subsec:exp_comparison}
To evaluate that the proposed CAAE can generate more photo-realistic results, we compare ours with the ground truth and the best results from prior works~\cite{wangrecurrent,kemelmacher2014illumination,shu2015personalized,suo2010compositional}, respectively.
We choose FGNET~\cite{lanitis2002toward} as the testing dataset, which has 1002 images of 82 subjects aging from 0 to 69.

\begin{figure}[h]
	\centering
	\includegraphics[width=.9\columnwidth]{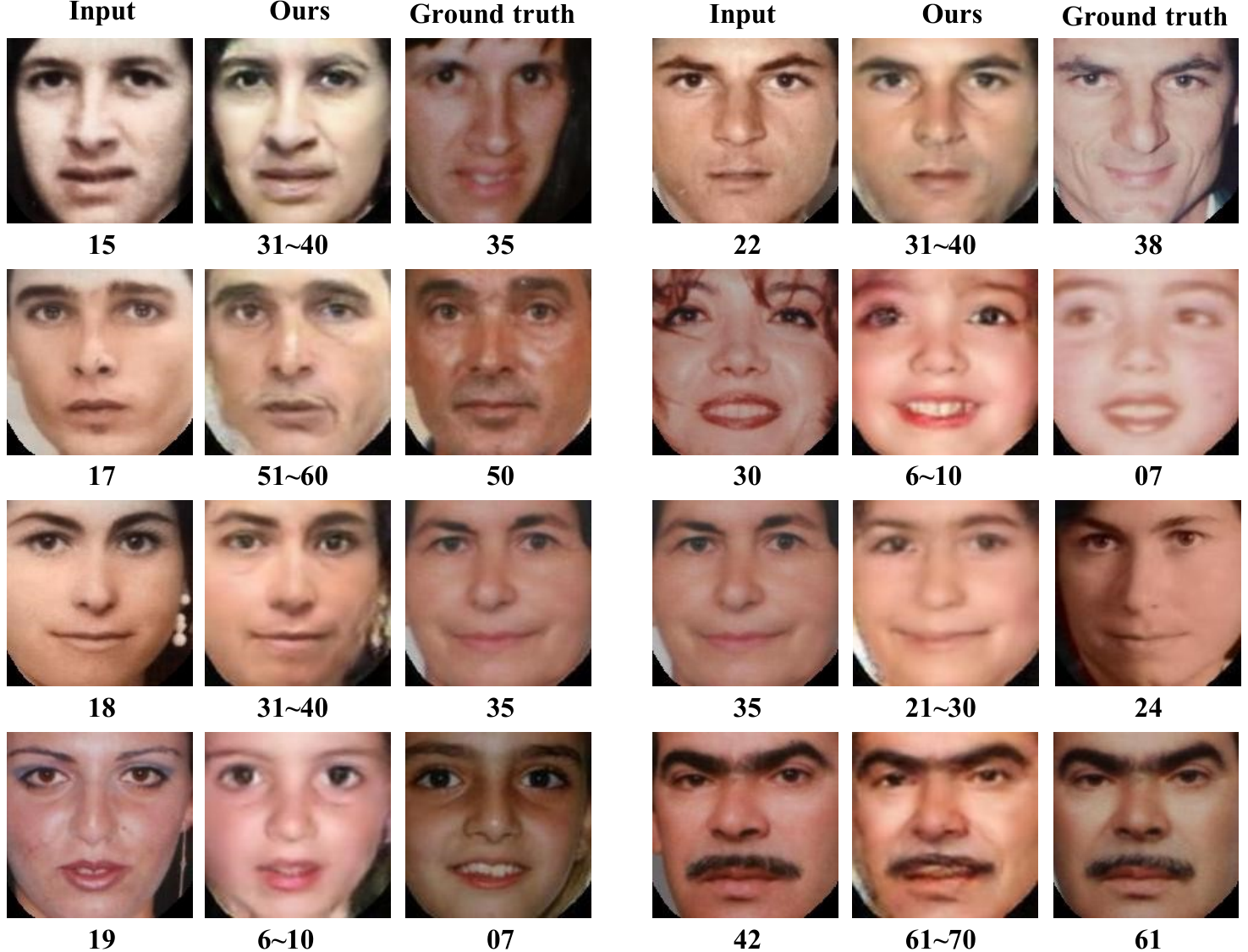}
	\caption{Comparison to the ground truth. }
	\label{fig:cmp2truth}
\end{figure}

\textbf{Comparison with ground truth}: In order to verify whether the personality has been preserved by the proposed CAAE, we qualitatively and quantitatively compare the generated faces with the ground truth. The qualitative comparison is shown in Fig.~\ref{fig:cmp2truth}, which shows appealing similarity. 
To quantitatively evaluate the performance, we pair the generated faces with the ground truth whose age gap is larger than 20 years. There are 856 pairs in total. We design a survey to compare the similarity where 63 volunteers participate. Each volunteer is presented with three images, an original image X, a generated image A, and the corresponding ground truth image B under the same group. They are asked whether the generated image A looks similar to the ground truth B; or not sure. We ask the volunteers to randomly choose 45 questions and leave the rest blank. We receive 3208 votes in total, with 48.38\% indicating that the generated image A is the same person as the ground truth, 29.58\% indicating they are not, and 22.04\% not sure. The voting results demonstrate that we can effectively generate photo-realistic faces under different ages while preserving their personality.

\begin{figure}[h]
	\centering
	\includegraphics[width=.8\columnwidth]{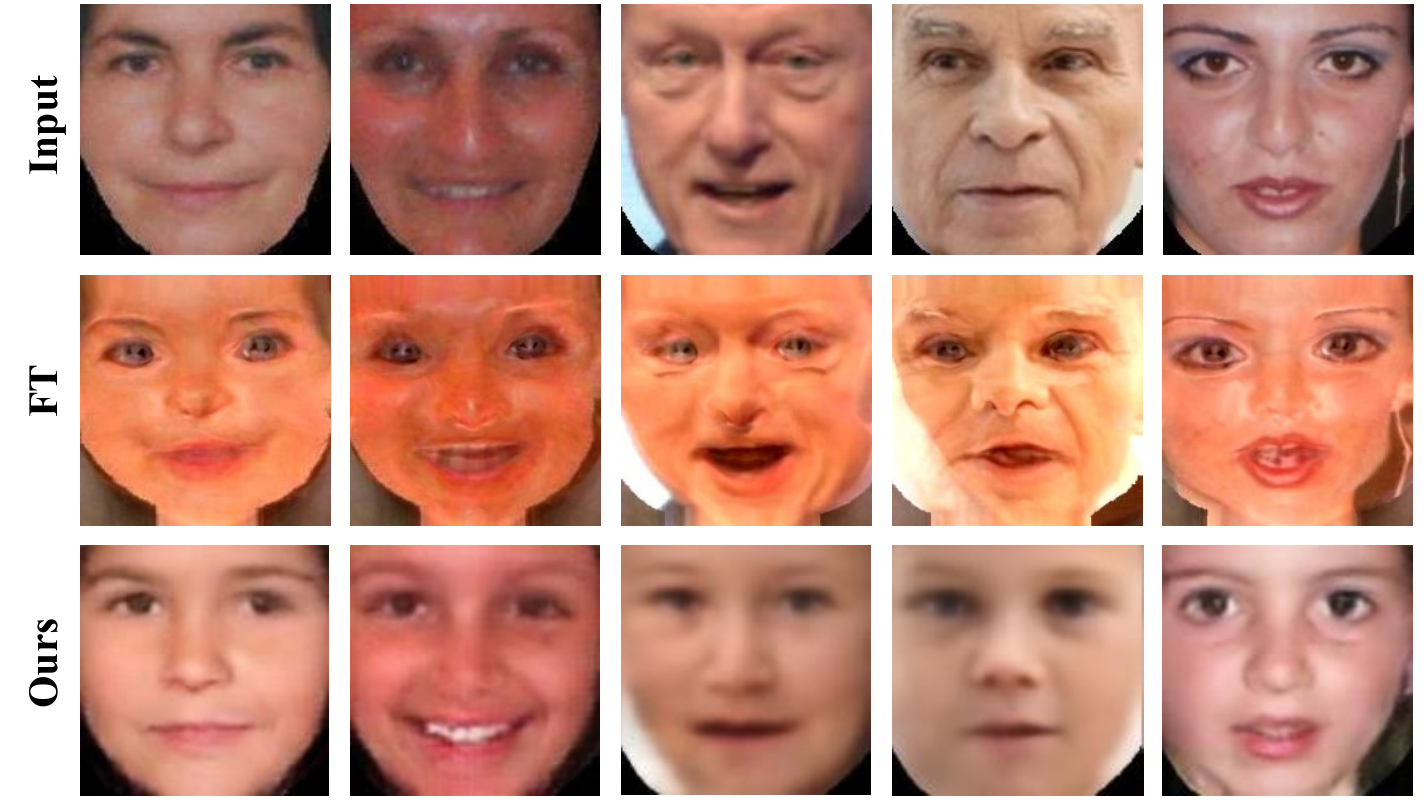}
	\caption{Comparison to prior work in rejuvenation. The first row shows the input faces, the middle row shows the baby faces generated by FT~\cite{FT} and the last row shows our results.}
	\label{fig:cmp2prior_regress}
\end{figure}

\textbf{Comparison with prior work}:
We compare the performance of our method with some prior works~\cite{wangrecurrent,kemelmacher2014illumination,shu2015personalized,suo2010compositional}, for face age progression and Face Transformer (FT)~\cite{FT} for face age regression. To demonstrate the advantages of CAAE, we use the same input images collected from those prior works and perform long age span progression. To compare with prior works, we cite their results as shown in Fig.~\ref{fig:cmp2prior}. We also compare with age regression works using the FT demo~\cite{FT} as shown in Fig.~\ref{fig:cmp2prior_regress}. Our results obviously show higher fidelity, demonstrating the capability of CAAE in achieving smooth face aging and rejuvenation. CAAE better preserves the personality even with a long age span. In addition, our results provide richer texture (\eg, wrinkle for old faces), making old faces look more realistic.
Another survey is conducted to statistically evaluate the performance as compared with prior works, where for each testing image, the volunteer is to select the better result from CAAE or prior works, or hard to tell. We collect 235 paired images of 79 subjects from previous works~\cite{wangrecurrent,kemelmacher2014illumination,shu2015personalized,suo2010compositional}. We receive 47 responses and 1508 votes in total with 52.77\% indicating CAAE is better, 28.99\% indicating the prior work is better, and 18.24\% indicating they are equal. This result further verifies the superior performance of the proposed CAAE. 
\subsection{Tolerance to Pose, Expression, and Occlusion}
\label{subsec:exp_tolerance}
\begin{figure}[h]
	\centering
	\includegraphics[width=\columnwidth]{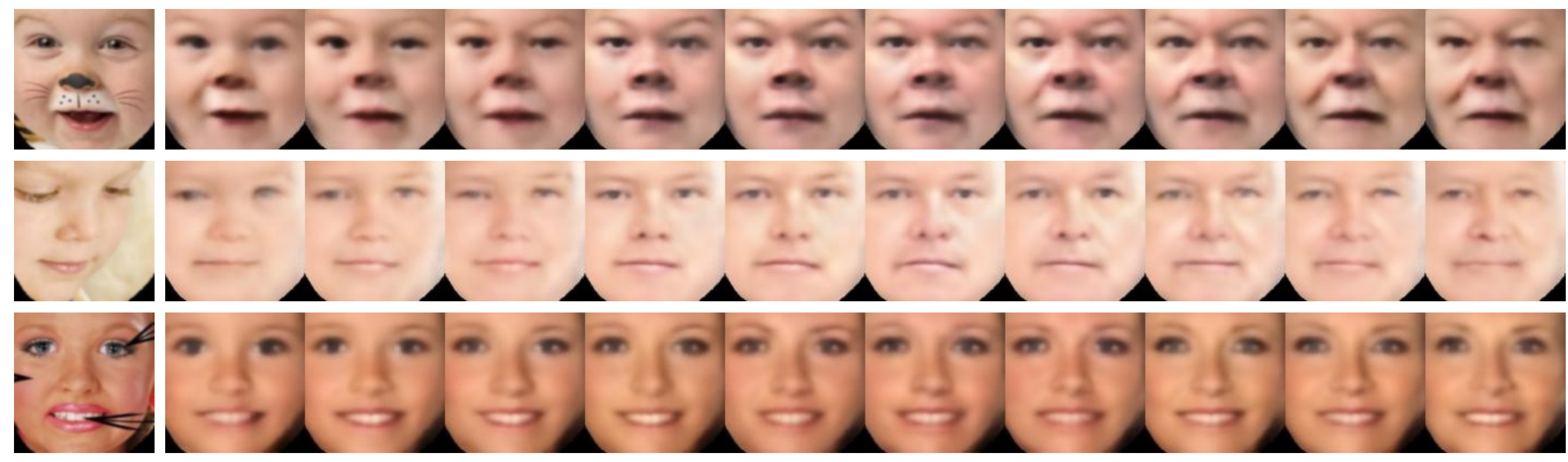}
	\caption{Tolerance to occlusion and variation in pose and expression. The very left column shows the input faces, and the right columns are generated faces by CAAE from younger to older ages. The first input face presents relatively more dramatic expression, the second input shows only the face profile, and the last one is partially occluded by facial marks.}
	\label{fig:tolerance}
\end{figure}

As mentioned above, the input images have large variation in pose, expression, and occlusion. To demonstrate the robustness of CAAE, we choose the faces with expression variation, non-frontal pose, and occlusion, respectively, as shown in Fig.~\ref{fig:tolerance}. It is worth noting that the previous works~\cite{wangrecurrent,kemelmacher2014illumination} often apply face normalization to alleviate from the variation of pose and expression but they may still suffer from the occlusion issue. In contrast, the proposed CAAE obtains the generated faces without the need to remove these variations, paving the way to robust performance in real applications. 

\section{Discussion and Future Works}
\label{sec:discussion}
In this paper, we proposed a novel conditional adversarial autoencoder (CAAE), which first achieves face age progression and regression in a holistic framework. We deviated from the conventional routine of group-based training by learning a manifold, making the aging progression/regression more flexible and manipulatable --- from an arbitrary query face without knowing its true age, we can freely produce faces at different ages, while at the same time preserving the personality. We demonstrated that with two discriminators imposed on the generator and encoder, respectively, the framework generates more photo-realistic faces. Flexibility, effectiveness, and robustness of CAAE have been demonstrated through extensive evaluation.  

The proposed framework has great potential to serve as a general framework for face-age related tasks. More specifically, we trained four sub-networks, \ie, $E$, $G$, $D_z$, and $D_{img}$, but only $E$ and $G$ are utilized in the testing stage. The $D_{img}$ is trained conditional on age. Therefore, it is able to tell whether the given face corresponds to a certain age, which is exactly the task of age estimation. For the encoder $E$, it maps faces to a latent vector (face feature), which preserves the personality regardless of age. Therefore, $E$ could be considered a candidate for cross-age recognition. The proposed framework could be easily applied to other image generation tasks, where the characteristics of the generated image can be controlled by the conditional label. In the future, we would extend current work to be a general framework, simultaneously achieving age progressing ($E$ and $G$), cross-age recognition ($E$), face morphing ($G$), and age estimation ($D_{img}$).



{\small
	\bibliographystyle{ieee}
	\bibliography{references}
}

\end{document}